\documentclass[conference]{IEEEtran}
\IEEEoverridecommandlockouts
\usepackage{cite}
\usepackage{amsmath,amssymb,amsfonts}
\usepackage{algorithmic}
\usepackage{graphicx}
\usepackage{textcomp}
\usepackage{xcolor}
\usepackage{times}
\usepackage{soul}
\usepackage{url}
\usepackage[hidelinks]{hyperref}
\usepackage[utf8]{inputenc}
\usepackage[small]{caption}
\usepackage{graphicx}
\usepackage{subfig}
\usepackage{amsmath}
\usepackage{booktabs}
\usepackage{algorithm}
\usepackage{algorithmic}
\usepackage{amsmath,amssymb,amsfonts}

\def\BibTeX{{\rm B\kern-.05em{\sc i\kern-.025em b}\kern-.08em
		T\kern-.1667em\lower.7ex\hbox{E}\kern-.125emX}}
\begin{document}
	
	\title{Inpatient2Vec: Medical Representation Learning \\ for Inpatients}
	
	%\author{
	%	\IEEEauthorblockN{Ying Wang\IEEEauthorrefmark{1}}
	%	\IEEEauthorblockA{\textit{School of Software, Tsinghua University} \\
	%		Beijing, China \\
	%			wangyin-17@mails.tsinghua.edu.cn}
	%	\and
	%	\IEEEauthorblockN{Xiao Xu\IEEEauthorrefmark{1}}
	%	\IEEEauthorblockA{\textit{Ping An Health Technology} \\
	%		Beijing, China \\
	%			xuxiao780@pingan.com.cn,}
	%	\and
	%	\IEEEauthorblockN{Tao Jin\IEEEauthorrefmark{2}}
	%	\IEEEauthorblockA{\textit{School of Software, Tsinghua University} \\
	%		Beijing, China \\
	%			jintao05@gmail.com}
	%	\and
	%	\IEEEauthorblockN{Xiang Li}
	%	\IEEEauthorblockA{\textit{Ping An Health Technology} \\
	%		Beijing, China \\
	%			lixiang453@pingan.com.cn}
	%	\and
	%	\IEEEauthorblockN{Guotong Xie}
	%	\IEEEauthorblockA{\textit{Ping An Health Technology} \\
	%		Beijing, China \\
	%			xieguotong@pingan.com.cn}
	%	\and
	%	\IEEEauthorblockN{jianmin Wang}
	%	\IEEEauthorblockA{\textit{School of Software, Tsinghua University} \\
	%		Beijing, China \\
	%		jimwang@tsinghua.edu.cn}
	%		
	%	\thanks{\IEEEauthorrefmark{1} Equal Contribution.}
	%	\thanks{\IEEEauthorrefmark{2} Corresponding author.}
	%}
	
	\author{\IEEEauthorblockN{Ying Wang\IEEEauthorrefmark{3}\IEEEauthorrefmark{1}, Xiao Xu\IEEEauthorrefmark{2}\IEEEauthorrefmark{1},
			Tao Jin\IEEEauthorrefmark{3}\IEEEauthorrefmark{4}, Xiang Li\IEEEauthorrefmark{2}, Guotong Xie\IEEEauthorrefmark{2}, Jianmin Wang\IEEEauthorrefmark{3}}
		\IEEEauthorblockA{\IEEEauthorrefmark{3}School of Software, Tsinghua University\\
			\IEEEauthorrefmark{2}Ping An Health Technology\\
			Email: wang-yin17@mails.tsinghua.edu.cn, xuxiao780@pingan.com.cn,\\ jintao05@gmail.com, lixiang453@pingan.com.cn, \\
			xieguotong@pingan.com.cn, jimwang@tsinghua.edu.cn}
		\thanks{\IEEEauthorrefmark{1} Equal Contribution.}
		\thanks{\IEEEauthorrefmark{4} Corresponding author.}
	}

	\maketitle
	
	\begin{abstract}
		Representation learning (RL) plays an important role in extracting proper representations from complex medical data for various analyzing tasks, such as patient grouping, clinical endpoint prediction and medication recommendation. Medical data can be divided into two typical categories, outpatient and inpatient, that have different data characteristics. However, few of existing RL methods are specially designed for inpatients data, which have strong temporal relations and consistent diagnosis. In addition, for unordered medical activity set, existing medical RL methods utilize a simple pooling strategy, which would result in indistinguishable contributions among the activities for learning. In this work, we propose Inpatient2Vec, a novel model for learning three kinds of representations for inpatient, including medical activity, hospital day and diagnosis. A multi-layer self-attention mechanism with two training tasks is designed to capture the inpatient data characteristics and process the unordered set. Using a real-world dataset, we demonstrate that the proposed approach outperforms the competitive baselines on semantic similarity measurement and clinical events prediction tasks. 
	\end{abstract}
	
	
	\section{Introduction}
	Inpatient is an important clinical scenario which has accumulated massive data. A lot of machine learning methods have been proposed to tackle the analysis tasks for inpatient data, such as predicting in-hospital mortality\cite{tabak2013using}, readmission\cite{nguyen2016predicting}, next day activity\cite{xu2018learning} and length-of-stay (LOS)\cite{liu2010length, barnes2015real}. The quality of data representation heavily determines the performance of these methods \cite{bengio2013representation}.
	
\begin{figure}[htb]
		\centering
		\subfloat[A patient's multiple visits (journey) with different diagnosis and activities. ]{
			\label{Inter-intra:out}
			\includegraphics[width=1\columnwidth]{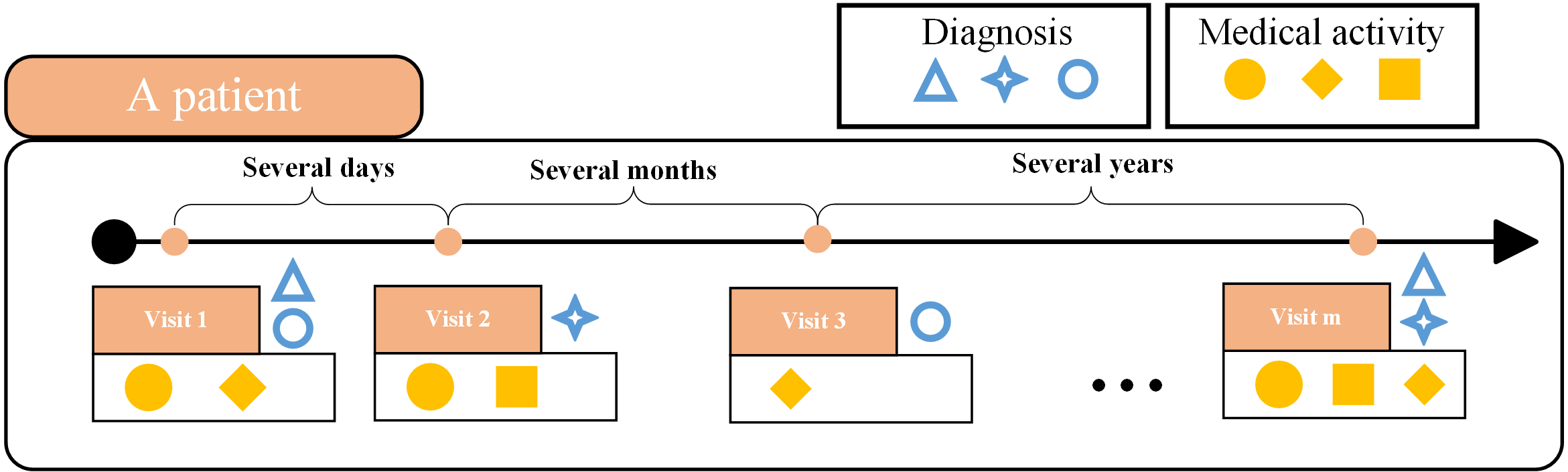}}\\
		
		\subfloat[Multiple days with consistent diagnosis and different activities in an inpatient visit. ]{
			\label{Inter-intra:in}
			\includegraphics[width=1\columnwidth]{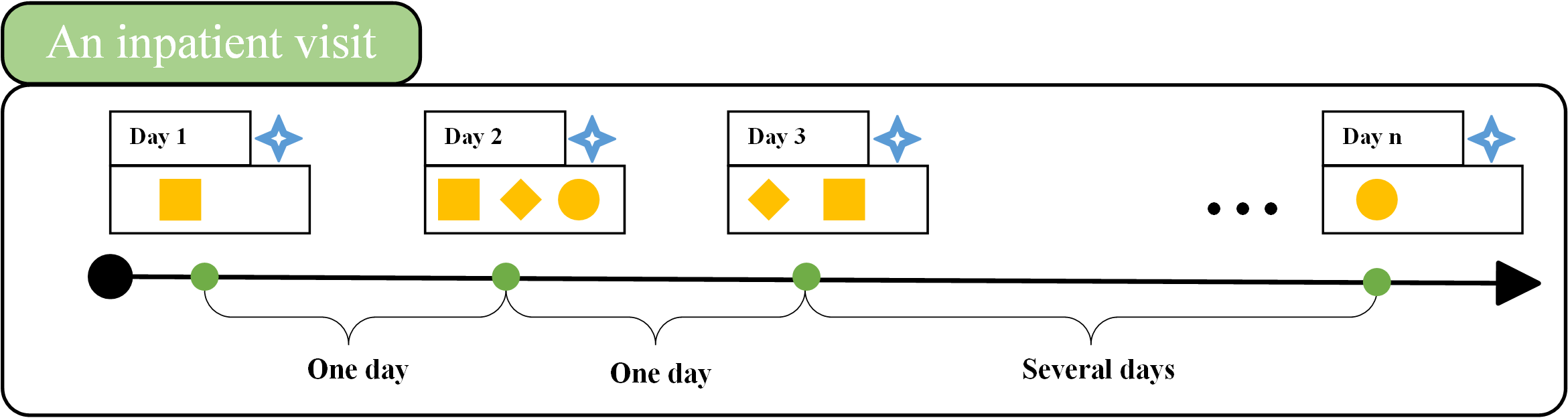}}
		\caption{An example for different medical data forms. }
		\label{Inter-intra}
		
	\end{figure}

	The remarkable success of deep learning technologies in a wide range of complex tasks spanning from computer vision, question answering and machine translation, is of great capacity in this area \cite{miotto2017deep,ravi2017deep,shickel2017deep}. High-quality distributed representations for various medical concepts, such as diagnosis, medical activities (drugs and procedures), hospital visits and patients' journeys, can be extracted end-to-end without human intervene \cite{choi2016learning,choi2016multi,choi2016medical,liu2018learning,cai2018medical,choi2017gram}. With the outstanding ability, a variety of studies achieved excellent performance on different clinical tasks \cite{choi2016retain,cheng2016risk,zhu2016measuring,baytas2017patient,ma2017dipole}. However, most of them focued on outpatient data, rather than inpatient data. Fig.\ref{Inter-intra:out} shows a typical outpatient data form that these methods applied on. The medical activities in a visit are unordered, while the patient's visits are ordered. They use the co-occurrence information and temporal relations in such data to construct the deep neural networks for RL, following the key principle of word2vec \cite{mikolov2013distributed} that similar words (medical concepts) share similar context. 
	
	While inpatient data\footnote{Intensive-Care-Unit (ICU) patients is not considered in this paper. } have distinct form and analyzing goals compared to the above one. As illustrated in Fig.\ref{Inter-intra:in}, an inpatient visit is composed by several temporal related days, and the medical activities in a day are unordered. For inpatient data, existing medical RL methods are facing following challenges: 
	
	\paragraph{Temporal relations} Temporal feature plays a vital role in RL for medical data. For inpatient, the temporal relation is reflected in \textit{day-level}, which are stronger than the \textit{visit-level} relation of outpatient. For example, there are two consecutive visits of an outpatient that the pre-visit is for common cold and the post-visit is for fracture. Even if their time interval is short (e.g. several days), the temporal relation between the two visits is weak. In contrast, the treatment for most of inpatients is on a daily basis, so that the days in sequence are closely related. Therefore, the stronger temporal relations should be taken into account for the inpatient RL.

	\paragraph{Importance of diagnosis} In most of previous RL methods on medical data, diagnosis is usually treated as a kind of medical activities \cite{choi2016multi,cai2018medical}, because a patient's multiple visits may correspond to different diagnose information. It means that diagnosis would be mapped to the same representation space as medical activities. While for inpatient, diagnosis plays a guidance role for all the days \cite{xu2018learning}. In this work, we highlight the importance of the first diagnosis of each inpatient visit for RL. 
	
\begin{figure}[htb]
		\centering
		\subfloat[To learn the representation of $i$-th medical activity in $t$-th unordered set, pooling operation is applied on context unordered set ($\{t-2,t-1,t,t+1,t+2\}$, without the target activity). $h_{t}^{i}$ is the aggregation of the pooling results. ]{
			\label{other:a}
			\includegraphics[width=1\columnwidth]{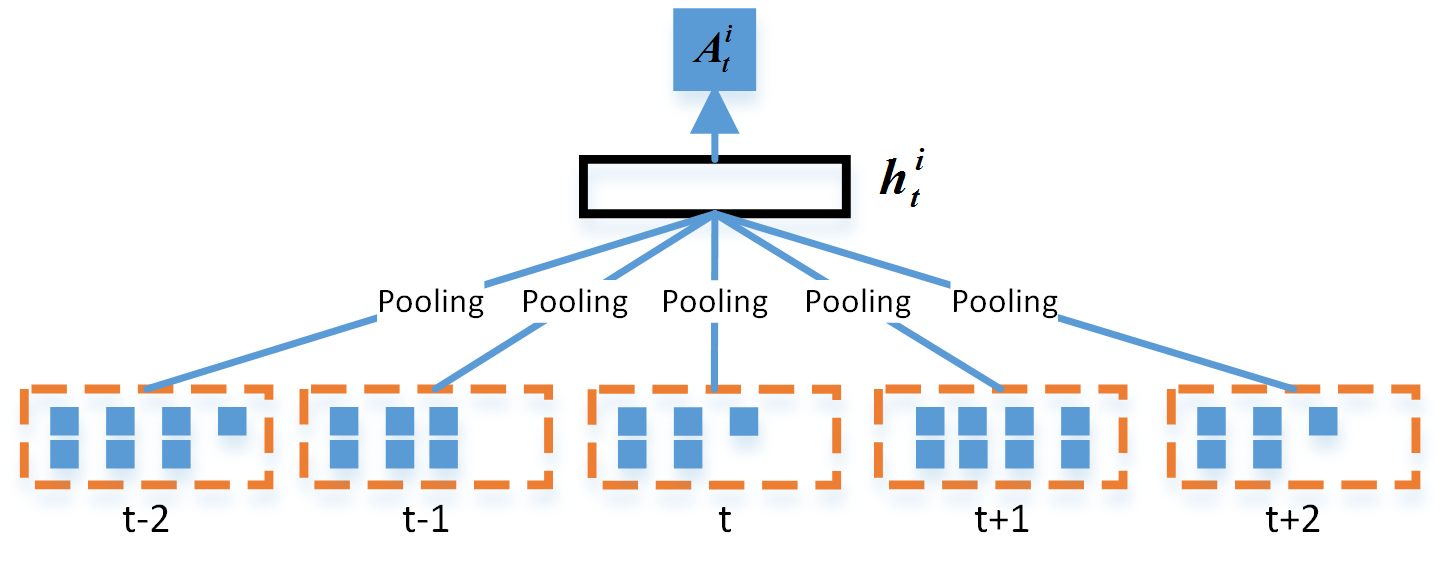}}\\
		
		\subfloat[To learn the representation of $t$-th unordered set, pooling operation is applied on context unordered set ($\{t-2,t-1,t+1,t+2\}$). ]{
			\label{other:v}
			\includegraphics[width=1\columnwidth]{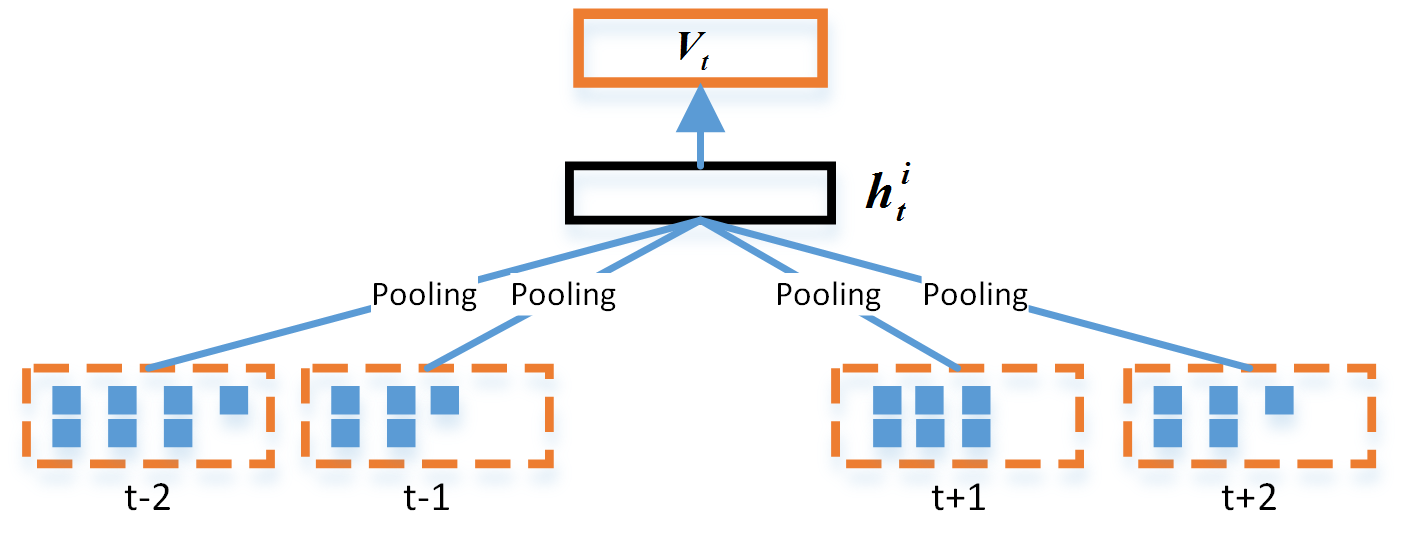}}
		\caption{Pooling strategy for unordered medical activity set. }
		\label{other}
	\end{figure}

	\paragraph{Unordered medical activity set} As mentioned before, there are medical activities with same time-stamps in medical data. It is different from the natural language area that words in a sentence are always in sequence. To solve this problem, a popular strategy is the use of a pooling operation on the unordered set, such as sum, average and maximum, to generate the medical activity representation \cite{cai2018medical,choi2016learning} (see Fig.\ref{other:a}) or visit representation \cite{choi2016multi} (see Fig.\ref{other:v}). However, pooling makes each medical activity in the unordered set have equal contribution to the RL, which is not in conformity with the clinical practice. 
	
	To tackle the above three challenges, this paper propose \textbf{Inpatient2Vec}, a novel medical RL approach for inpatients. We aim to learn three kinds of representations: (medical) activity, (hospital) day and diagnosis, which can not only cover the core data characteristics of inpatients, but also satisfy the requirements for various analyzing applications of inpatients. Inpatient2Vec is an extension of Bidirectional Encoder Representations from Transformers (BERT) \cite{devlin2018bert} that contains two learning tasks. The first one is \textit{masked activity prediction} which uses a percentage of activities in a day to predict the other activities in the day. Co-occurrence in the unordered set is utilized in this task. The second one is \textit{next day activity prediction} which uses the pre-days of an inpatient to predict the activities in the next day by a bi-directional LSTM \cite{hochreiter1997long}.This task highlights the temporal relations between ordered days. We design a Transformer-based network \cite{vaswani2017attention}, which combined activity, day and diagnosis representations, for the two learning tasks. In this network, the diagnosis plays a guidance role for the days through a time-aware mechanism, and the contributions of different activities in a day are distinguished by an attention mechanism. 
	
	The main contributions of this study are summarized bellow: 
	
	\begin{enumerate}
		\item Realizing the difference between inpatient and outpatient, we propose an effective RL approach for inpatients to cope with the three challenges, including temporal relations, importance of diagnosis and unordered medical activity set. A Transformer-based network is presented to capture the correlations between activity, day and diagnosis representations, and two learning tasks are designed as the training objective. 
		\item As a pre-training model for inpatients, the poposed method can be easily integrated into different clinical prediction models by fine-tuning. 
		\item We conduct experiments on real-world data to evaluate the quality of the learned representations from two aspects, that one is the semantic similarity and the other one is the performance for prediction tasks. The results shows that Inpatient2Vec outperforms the competing baselines. 
	\end{enumerate}
	
	\section{Related Works}
	Representing the massive and complex data in low-dimensional dense vectors is one of the core advantages of deep learning technologies. Existing works can be divided into two categories: one focuses on directly learning the representations of various medical concepts, and the other one utilizes end-to-end frameworks which can generate proper representations for different healthcare tasks. 
	
	\subsection{Representation learning for medical concepts}
	In \cite{choi2016learning}, word2vec was used to generate the embedding of activities, which were treated as the words in sentence. For the unordered activities in a visit, a random shuffle strategy was used to generate a sequence to satisfy the requirement of word2vec. Med2vec \cite{choi2016multi} derived visit and activity embedding by a multi-layer representation learning framework. A visiting embedding was generated by a sum pooling of the activities in the visit. In \cite{cai2018medical}, an attention mechanism was introduced into the continuous bag-of-words model (CBOW). GRAM \cite{choi2017gram} proposed a graph-based attention model to combine the information in EHR data and medical ontologies to learn the representation of medical concepts. MiME \cite{choi2018mime} solved a problem in outpatient data that there are multiple diagnosis for one visit and each diagnosis may correspond to a set of medical activities in the visit.

	\subsection{Representation learning for healthcare tasks}
	Retain \cite{choi2016retain} presented a two-level attention on reversed RNN for heart failure prediction. In  \cite{cheng2016risk}, an one-dimensional CNN with fusion mechanism was utilized for prediction chronic disease onset, including congestive heart failure and chronic obstructive pulmonary disease.  \cite{baytas2017patient} modified the gate structure of LSTM to capture the time irregularity between visits. A prediction and subtyping task for parkinson's disease was used to train the network. Dipole \cite{ma2017dipole} proposed three types of attention mechanisms on bi-directional RNN for diagnosis prediction. DPMI \cite{xu2018a} presented a deep predictive model for inpatients which address the challenge about fixed diagnosis and time irregularity. Patient2Vec \cite{zhang2018patient2vec} adopted a prediction task to get the patient representation. The input was generated by word2vec technology. In the works of this paragraph, the representations of medical concepts are generated from the end-to-end pipelines of different tasks. However, it is hard to transfer these representations for others tasks. 
	
	\section{Methodology}
	In this section, we give some related notations and definitions firstly. Then we give a brief introduction of BERT, which is the inspiration of our proposed method Inpatient2Vec. Finally, we show the details of Inpatient2Vec, including the input sequence construction and two kinds of unsupervised training tasks for RL. 
	
	\subsection{Preliminaries}
	We denote the set of inpatient visits as $\mathcal{V} = \{V_{1}, V_{2}, \cdots, V_{|\mathcal{V}|}\}$, where $|\mathcal{V}|$ is the number of visits in our dataset. Each inpatient visit may have more than one diagnosis, while in this paper we only concern the first diagnosis which largely determines the treatment strategy during the visit. The diagnosis code is represent as $\mathcal{G} = \{g_{1}, g_{2}, \cdots, g_{|\mathcal{G}|}\}$ with size $|\mathcal{G}|$. For an inpatient visit $V_{i}$, its diagnosis is denoted as $g^{(i)} \in \mathcal{G}$. Each inpatient visit is composed by several days, $V_{i} = \{d_{i,1}, d_{i,2}, \cdots, d_{i,n} \}$ where $n$ is the number of days in $V_{i}$. Each day contains a set of medical activities (e.g. drugs, procedures and nursing cares), and we defined all the unique activities in our dataset as $\mathcal{A} = \{a_{1}, a_{2}, \cdots, a_{|\mathcal{A}|}\}$ with size $|\mathcal{A}|$. 
	
	\subsection{Brief introduction of BERT}
	BERT \cite{devlin2018bert} achieved the state-of-the-art performance on a series of natural language processing tasks. It based on a multi-layer Transformer \cite{vaswani2017attention}, a widely used feature extractor. BERT consists of two parts: pre-training that considers both left and right context, and fine-tuning. In pre-training stage, two unsupervised tasks, masked language model and next sentence prediction, are used to generate the representations of each word and sentence. In fine-tuning stage, the generated representations are fed into the downstream tasks. 
	
	\subsection{Model}
	The simplest way that using BERT on inpatient data is to map the concept day and activity of inpatient to the sentence and word of BERT, respectively. However, there exists two weaknesses: 
	\begin{itemize}
		\item Diagnosis, a crucial concept for inpatients, can not be simply mapped to any elements in BERT. 
		\item BERT used a pair-wise sentence prediction task to catch sentence-level relations. However, it ignores the stronger long-term temporality in days of inpatients, which can not be utilized by this task. 
	\end{itemize}
	
	To solve the problems, Inpatient2Vec is an extension of BERT, which contains three kinds of representations, including activity, day and diagnosis. The core component of Inpatient2Vec is Transformer, a day-level feature extractor. We construct a sequence that is composed by the activities and diagnosis in a day as the input for Transformer. Two domain-related tasks are designed to learn the representations. 
	
	\subsubsection{Input sequence construction}\label{sec:isc}
\begin{figure}[htb]
		\centering
		\includegraphics[width=1.0\linewidth]{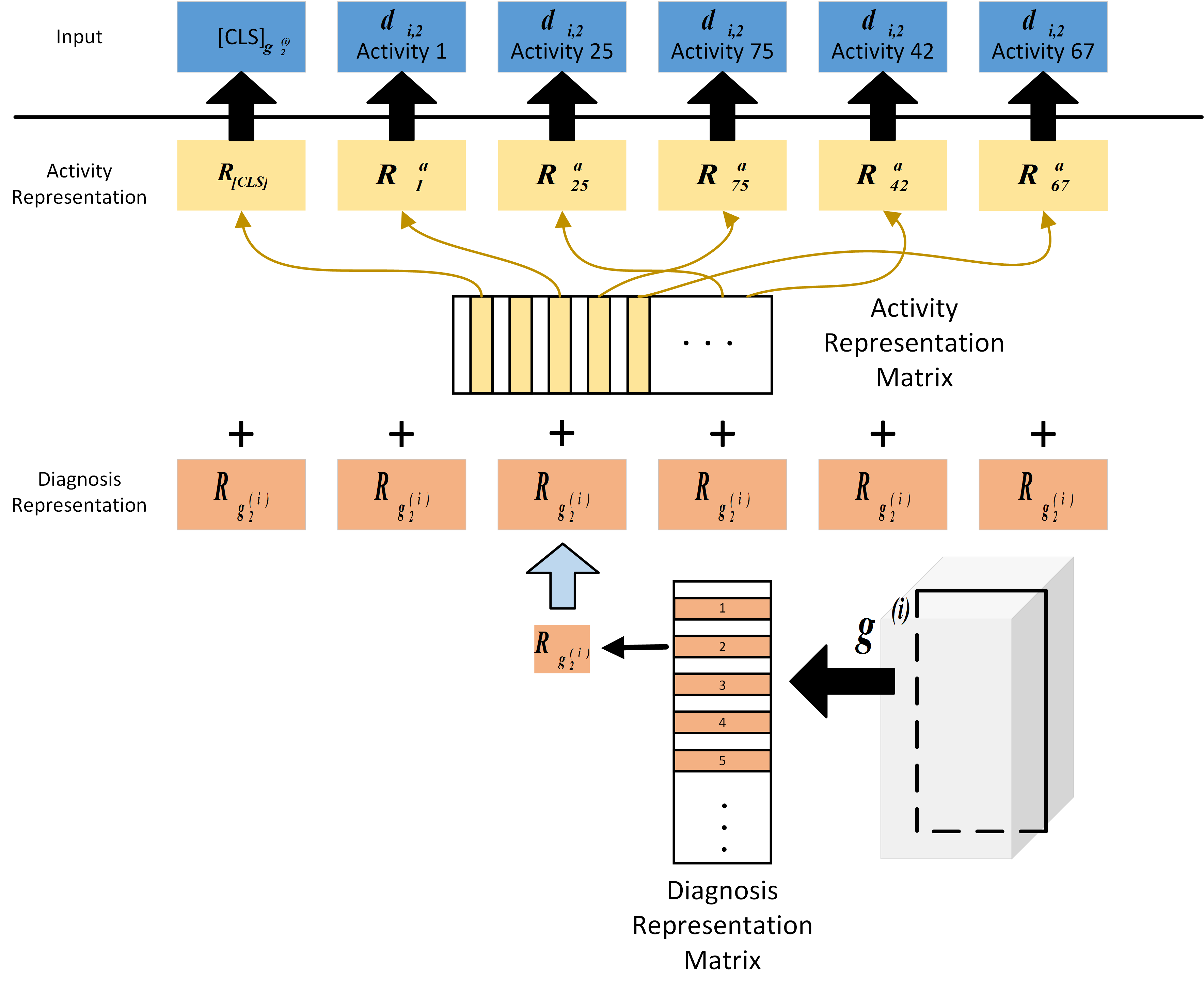}
		\caption{The construction of input sequence. }
		\label{input-representation}
	\end{figure}

	We firstly conduct a concept mapping between natural language and inpatient data. In natural language, a document is composed by a set of sentences, and each sentence contains several words. Analogically, we treat inpatient visit, day and activity as document, sentence and word, respectively. Similar to the construction of input sequence for Transformer in BERT, we combine the representations of the activities and diagnosis of a day as our input sequence (see the blue rectangles in Fig.\ref{input-representation}). The order of the activities in sequence can be arbitrary, because Transformer is insensitive to the order information. It meets the fact that all the activities in a day are an unordered set. Note that the first item ([CLS]$_{g_{2}^{(i)}}$) of the sequence is a special day-element which would be used for generating day representation through Transformer (we will detail it in the next section). 
	
	\begin{itemize}
		\item \textbf{Activity representation}. We map each medical activity to a low-dimension representation with size $q$ (see the yellow rectangles in Fig.\ref{input-representation}), which is denoted as $R^a_{i} \in \mathbb{R}^{q}$. Specially, we create an activity representation $R^a_{[CLS]}$ for [CLS]. 
		
		\item \textbf{Diagnosis representation}. For an inpatient visit, the diagnosis plays a guidance role in all the days. Considering the treatment for a diagnosis is usually based on days, we map each diagnosis $g_{i}$ to a matrix, $R_{g_{i}} \in \mathbb{R}^{N_{g_{i}} \times p}$, where $N_{g_{i}}$ is the maximum LOS of inpatient visits with diagnosis $g_{i}$ (see the orange rectangles in Fig.\ref{input-representation}). There are two advantages to adopt the three-dimension diagnosis representation. The first one is that the core treatment information in each day of one diagnosis is preserved in a vector with size $p$. The second one is that the actual time information have been involved in the RL architecture, which is similar to the positional embedding in BERT \cite{devlin2018bert}. 
	\end{itemize}
	
	Therefore, each item in the input sequence is the aggregation of the corresponded activity (yellow rectangles) and diagnosis (orange rectangles) representation. 
	
	\subsubsection{Unsupervised training tasks}
	In this part, we introduce the two unsupervised training tasks used to learn the inpatient representations. 
	
	\begin{itemize}
		\item \textbf{Masked activity prediction}. In this task, we want to utilize the co-occurrence among medical activities in a day to train the RL model. The activities frequently occurred together may refers to similar clinical function, such as the drugs for anti-inflammation and analgesia. We randomly mask 15\% activities in all days. A special activity representation $R_{[MASK]}$ is used to replace the masked activities, which is similar to $R_{[CLS]}$. Then, for each masked activity in a day, we use the other activities in the day to predict it. 
		
		\item \textbf{Next day activity prediction}. Previous task only considers context activities in a day, and the temporal relations between days have not been used. For inpatient visit, the treatment of a day strongly depends on the previous days, so we propose next day activity prediction task to capture the temporality. Specifically, given the previous days of an inpatient visit, the task is to predict the most likely activities that would be used in the next day. It is worth mentioning that this task is different from BERT (next sentence prediction), which only use the temporal relations between two consecutive sentences. Our task can better capture the long-term temporality among all the days in a visit. 
	\end{itemize}
	
	\subsubsection{Model architecture}
\begin{figure}[htb]
		\centering
		\includegraphics[width=0.9\columnwidth]{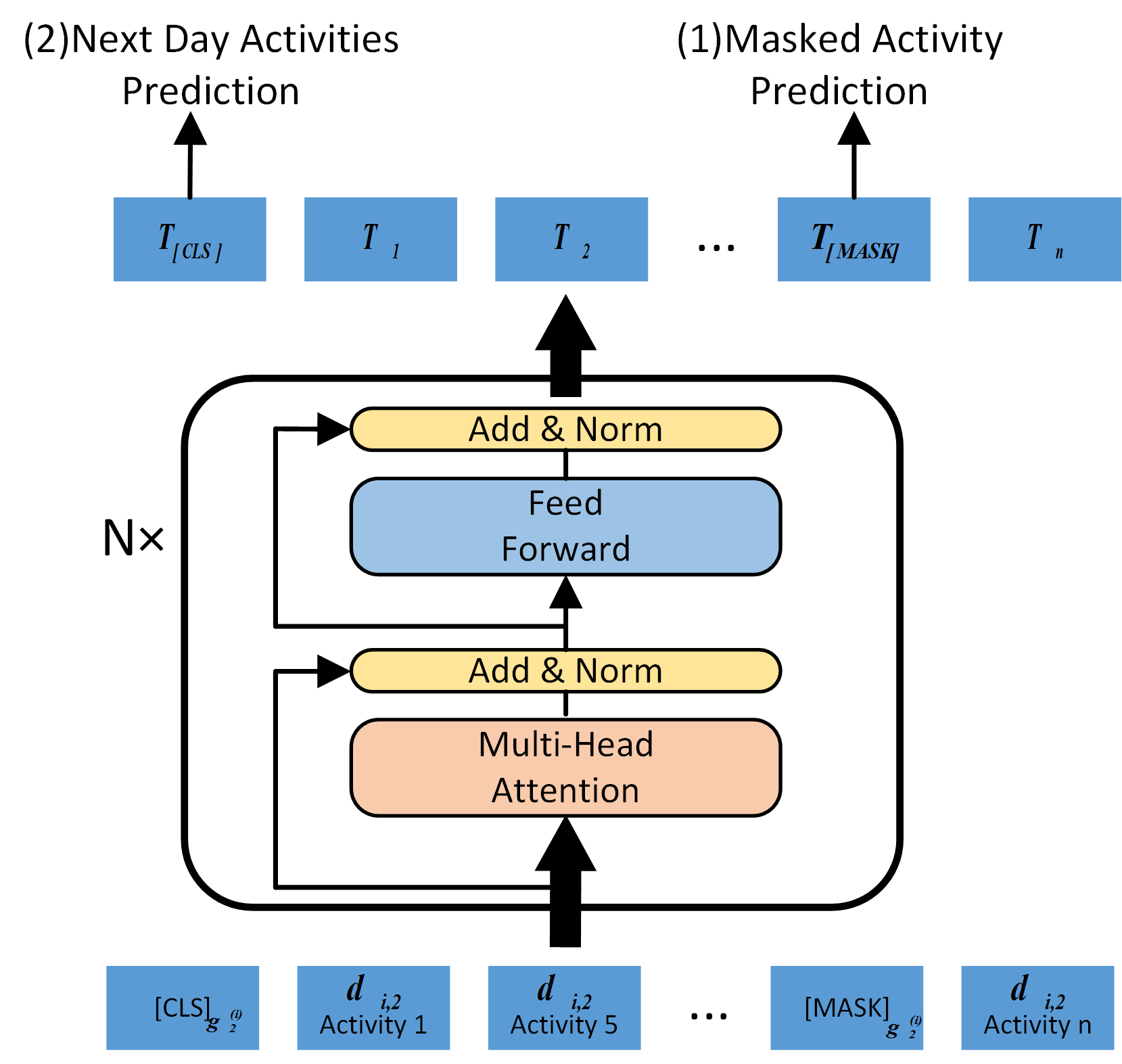}
		\caption{The model architecture of Inpatient2Vec. }
		\label{pre-training}
	\end{figure}

	Fig.\ref{pre-training} shows the architecture of our model. The bottom rectangles are the input sequence generated from a day. We fed the input sequence into a N-layer Transformer, which has been proved as an outstanding feature extractor. Each layer is consisted of a multi-head attention, a feed forward and two normalization layers. Equation (\ref{equ:self-attetion}), which is composed of three matrices: \textbf{Q}uery, \textbf{K}ey and \textbf{V}alue, is a portion of multi-head attention. 
	
	\begin{equation}
	\label{equ:self-attetion}
	Attention(Q,K,V) = Softmax(\frac{QK^{T}}{\sqrt{d_{k}}})V
	\end{equation}
	
	In our model, Q, K and V are set to equal, which is called self-attention. It can figure out the attentions among activities that represent the different importances for RL.  
	
	Through the Transformer, we can get the day representations as described as follows.  
	
	\begin{itemize}
		\item \textbf{Day representation}. A day representation $T_{[CLS]}$ is generated from the day-element ([CLS]$_{g_{t}^{(i)}}$). With the help of Transformer, the day representation $T_{[CLS]}$ takes into account the relations with activities in the day.  
	\end{itemize}
	
	In addition, besides $T_{[CLS]}$, the other outputs of Transformer (top rectangles in Fig.\ref{pre-training}) can be regarded as the day-based activity representations. This representation reflects the activity ambiguity in different days. The two unsupervised training tasks (masked activity prediction and next day activity prediction) take $T_{[MASK]}$ and $T_{[CLS]}$ as the input, respectively. 
	
	For masked activity prediction task, the prediction result is defined as follows. 
	
	\begin{equation}
	\hat{y}^{mask} = Softmax( T_{[MASK]}W_{a} + b_{a} )
	\end{equation}
	
	\noindent
	we calculate the loss $\mathcal{L}_{mask}$ according to the cross entropy between $\hat{y}^{mask}$ and the true label $y^{mask}$, which is the one-hot vector of the masked activity. 
	
	\begin{equation}
	\mathcal{L}_{mask} = -\frac{1}{M}\sum_{m=1}^{M}( y^{mask}_{m}\mbox{log}\hat{y}^{mask}_{m} + (1-y^{mask}_{m})\mbox{log}\hat{y}^{mask}_{m} )
	\end{equation}
	
	\noindent
	where $M$ is the number of masked activities in all visits. 
	
	For next day activity prediction, we put the day representation into a single layer bi-directional LSTM. Previous t-1 days (from $d_{i,1}$ to $d_{i,t-1}$) are used to predict the activities may be used in the next day $d_{i,t}$. The hidden state $h_{i,t}$ of bi-directional LSTM is used for the prediction as follows. 
	
	\begin{equation}
	\hat{y}_{i,t} = Softmax( h_{i,t}W + b )
	\end{equation}
	
	\noindent
	Similar to masked activities prediction, the loss of this task, denoted as $\mathcal{L}_{next}$, is also the cross entropy between $\hat{y}_{i,t}$ and true label $y_{i,t}$. 
	
	We add the $\mathcal{L}_{mask}$ and $\mathcal{L}_{next}$ as the loss function to train the model.

	\section{Experiment}
	In this section, experiments are conducted to demonstrate the effectiveness of the learned representations. We firstly give the description of dataset and experimental setting, including baseline methods, evaluation and implementation details. Then we analyze the experimental results. 
	
	\subsection{Dataset}
	We use a real-world insurance claims dataset which comes from a Chinese city. We filter out some data according to following criteria: (1) a visit whose the number of days is less than 2 or more than 50; (2) a diagnosis whose number of visits is out of range 100 to 3000. Each visit corresponds to a diagnosis code that follows the ICD-10. Table \ref{tab:dataset statistic} lists the details about dataset.
	
	\begin{table}[htbp]
		\centering
		\caption{Statistics of our dataset}
		\label{tab:dataset statistic}
		{\small \begin{tabular}{ccc}
				\toprule[1.5pt]
				Items & Numbers \\ 
				\midrule[1.0pt]
				\# of visits & 226,420\\
				\# of days & 1,869,294 \\
				\# of diagnosis codes (ICD-10) & 479 \\
				\# of medical codes & 3,952 \\
				Avg. \# of activities per day & 13.97 \\
				Avg. of length of stay & 9.26\\
				\bottomrule[1.5pt]
		\end{tabular}}
	\end{table}
	
	\subsection{Experiment Setup}
	Through \textbf{Inpatient2Vec}, we can get three kinds of representations: activity, day and diagnosis representation. To comprehensively evaluate the performance of the learned representations, we designed two categories of experiments. The first one focuses on the semantic similarity, including the activity intrusion task for activity representations, and clustering task for diagnosis representations\footnote{We do not evaluate the semantic similarity for day representations because it is hard to find the corresponded ground truth. }. The second one contains two prediction tasks, next day activity prediction and remaining length-of-stay (LOS, refers to the number of days of an inpatient stay in hospital from admission to discharge) prediction, which are used to verify the applicability of the learned representation. Furthermore, we did an ablation study to evaluate the importances of different parts of Inpatient2Vec. 
	
	\subsubsection{Semantic similarity measurement}
	
	Inspired by the word intrusion task \cite{murphy2012learning,luo2015online} which is widely used in evaluating the semantic quality of word representations, we designed the activity intrusion task for activity representations. In this task, we firstly calculated the Euclidean distance between every two activity representations. Then, given an activity $A$, we constructed a set containing 6 activities that 5 of them are the top 5 nearest activities and the rest one (called "intrusion") comes from the last 50\% activities based on the Euclidean distance. Lastly, we invited three doctors to pick up the "intrusion" activity and calculated the precision of the correct picking as the measurement.
	
	For diagnosis representation, we used the hierarchy of ICD-10 as the clustering ground truth, that 479 diagnosis are grouped into 131 categories (keeping the top 3 characters). K-means (implemented by scikit-learn 0.19.0) was used as clustering method, and normalized mutual information(NMI) was used as the measurement. 
	
	We compared Inpatient2Vec against 3 state-of-the-art models, i.e, CBOW (as shown in Fig.\ref{other}), Med2Vec (based on skip-gram) \cite{choi2016multi} and RoMCP (an extension of CBOW for inpatients considering diagnosis information) \cite{xu2018learning}. The methods without generating day representations, such as the works in \cite{cai2018medical,choi2016learning}, are not considered for the comparison. It is worth mentioning that diagnosis is treated as a kind of activities in CBOW and Med2Vec, so that they would be mapped to the same representation space. 
	
	\subsubsection{Inpatient Prediction Tasks}
	
	The core goal of RL is to improve the performance of different analyzing tasks. In this part, we selected two typical inpatient prediction tasks for evaluation. One is next day activity prediction, which is same to the second training task of Inpatient2Vec. The other one is remaining LOS prediction, that calculate the possible number of days from each time-stamp to discharge. Three state-of-the-art prediction approaches were selected as the basic models. We evaluated that if the approaches could benefit from the pre-train representations from Med2Vec, RoMCP and Inpatient2Vec \footnote{CBOW is not considered here because of the similar architecture to Med2Vec.} on the two tasks through a fine-tuning procedure. 
	
	The three basic models are listed as follows: 1) \textbf{Retain} \cite{choi2016retain} is an interpretable predictive model with two kind of reverse time attention mechanism, which focus on visit level and day level. 2) \textbf{Dipole} \cite{ma2017dipole} is a bi-directional LSTM network, with three kinds of attention. We use the location-based attention, which performs best in our prediction tasks. 3) \textbf{T-LSTM} \cite{baytas2017patient} focuses on handling irregular time intervals in longitudinal patient records. 
	
	On the one hand, we evaluated the performance of the three models with original inputs (diagnosis and activities represented by one-hot vectors). On the other hand, we input the three pre-train representations to the models with fine-tuning for comparison. 
	
	For next day activity prediction, we calculate RECALL@k (sensitivity) for the correctly predicted medical activities in top k value of $\hat{y}_{i,t}$ as the measurement. 
	
	\begin{equation}
	Recall@k=\frac{|A_{i,t}|}{\mbox{min}(k,|d_{i,t}|)}, \ \ \ \  k \in \{5, 10, 20\}
	\end{equation}
	
	\noindent
	where $|A_{i,t}|$ is the activity count of the interaction between $d_{i,t}$ and the top k of $\hat{y}_{i,t}$, $|d_{i,t}|$ refer to the activity count that actually occurred in $d_{i,t}$. Besides, the variance of RECALL@k with adaptive k is also used as the measurement, which is defined as $RECALL@A = \frac{|A_{i,t}|}{|d_{i,t}|}$. Recall@K is a widely used evaluation measure in Clinical-Decision-Support-Systems, which can help doctors make a decision based on the topK recommendations. 
	
	For remaining LOS prediction, RMSE between the actual and predictive remaining LOS is used as the measurement. 
	
	\subsubsection{Implementation Details}
	All approaches are implemented in TensorFlow 1.12.0. We randomly divided dataset into the training, validation and testing set in a 0.75:0.1:0.15 ratio. The validation set is used to determine the values of hyper-parameters. For pre-training model, we use Adam with learning rate of 1e-4, $\beta_{1}$ = 0.9, $\beta_{2}$ = 0.999, L2 weight decay of 0.01. The diagnosis and activity representation size is 384, the number of attention head is 6, and the number of Transformer is 6. The hidden layer size of bi-directional LSTM is 200. For the two prediction tasks, we use Adadelta optimizer to train our model, with a mini-batch of 128 patients. The hidden layer size of Retain, Dipole and T-LSTM is 200. We execute 10 epochs and show the best performance for each approaches in above two tasks. 
	
	\subsection{Results Analysis}
	
	%\begin{figure}[htb]
	%	\centering
	%	\subfloat[Semantic similarity measurement. ]{
	%		\label{representation}
	%		\includegraphics[width=0.55\columnwidth]{pic/Representation_result}}
	%	\subfloat[Ablation study. ]{
	%		\label{abalation}
	%		\includegraphics[width=0.44\columnwidth]{pic/ablation}}\\	
	%	\caption{The results of semantic similarity measurement and ablation study. }
	%\end{figure}
	
	\begin{figure}[htb]
		\centering
		\includegraphics[width=0.8\columnwidth]{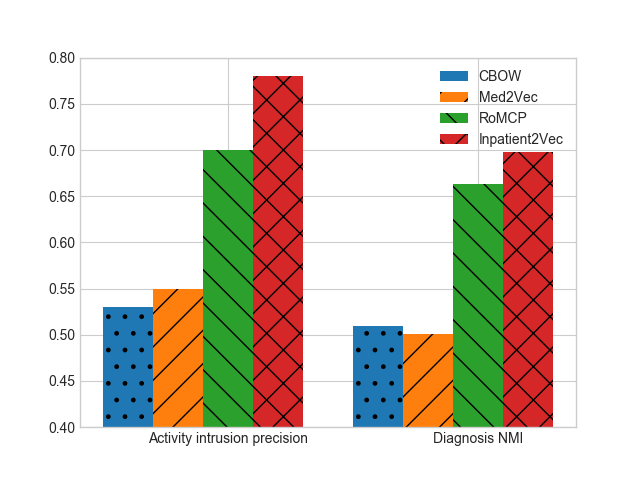}
		
		\caption{The results of semantic similarity measurement. }
		\label{representation_result}
	\end{figure}

	\begin{table*}[htp]
		
		\caption{The performance of the two predictive tasks for inpatient. }
		\label{tab:recall_results}
		
		{\small \begin{tabular*}{\linewidth}{@{\extracolsep{\fill}}cccccccccc}
				\toprule
				& RECALL@A & RECALL@5 & RECALL@10 & RECALL@20 & LOS\_RMSE \\ \midrule
				Retain               & 0.6823   & 0.4083   & 0.6635    & 0.8046    & 3.9817    \\
				Dipole               & 0.7104   & 0.4308   & 0.7094    & 0.8355    & 3.6583    \\
				T-LSTM               & 0.7131   & 0.4366   & 0.6937    & 0.8302    & 3.5018    \\ \midrule
				Med2Vec+Retain       & 0.6716   & 0.4099   & 0.6679    & 0.8054    & 3.9091    \\
				Med2Vec+Dipole       & 0.7119   & 0.4316   & 0.7093    & 0.8260    & 3.7028    \\
				Med2Vec+T-LSTM       & 0.7127   & 0.4397   & 0.6998    & 0.8337    & 3.5004    \\ \midrule
				RoMCP+Retain         & 0.7028   & 0.4144   & 0.6663    & 0.8125    & 3.7543    \\
				RoMCP+Dipole         & 0.7332   & 0.4407   & 0.7176    & 0.8308    & 3.4060    \\
				RoMCP+T-LSTM         & 0.7364   & 0.4376   & 0.7120    & 0.8358    & 3.4122    \\ \midrule
				Inpatient2Vec+Retain & 0.7231   & 0.4218   & 0.6766    & 0.8164    & 3.6780    \\
				Inpatient2Vec+Dipole & \textbf{0.7542}   & 0.4631   & \textbf{0.7303 }   & \textbf{0.8579}    & 3.3465    \\
				Inpatient2Vec+T-LSTM & 0.7537   & \textbf{0.4652}   & 0.7298    & 0.8507    & \textbf{3.3084}    \\ 
				\bottomrule
		\end{tabular*}}
	\end{table*}
	
	\paragraph{Activity intrusion} The precision of activity intrusion task among different RL methods is shown in the left side of Fig.\ref{representation_result}. We can observe that CBOW and Med2Vec, which adopt similar architectures, get the similar poor performances. RoMCP obtains nearly 40\% improvements compared to the CBOW and Med2Vec. The significant difference between them is that RoMCP considers the strict temporal relations in inpatient visit by concatenating the day representations for prediction. While in CBOW, the context days are simply aggregated to predict the center day, and in Med2Vec, the center day is used to predict all the context days. Inpatient2Vec outperforms the other methods. Compared to RoMCP, the improvement may stem from two aspects. One is that the usage of bi-directional LSTM can better capture the temporal relations with long distance. This further confirms that the stronger temporal relations play an important role in inpatient RL. The other one is that Transformer calculate the attention weights between activities in a day. In contrast to the pooling operation, our method distinguish the contributions between different activities. 
	
	\paragraph{Diagnosis clustering} The right side of Fig.\ref{representation_result} illustrates the NMI results for diagnosis clustering. It is observed that Inpatient2Vec and RoMCP perform better than CBOW and Med2Vec. The reason is that the latter two methods map diagnosis and activity into the same representation space, without considering the guidance role of diagnosis for inpatient. While in the former two methods, diagnosis information is regarded as an independent representation, which reserves the most important factors for the diagnosis. Inpatient2Vec shows sight improvement to RoMCP. The main reason is that the diagnosis representation in Inpatient2Vec are day-based, that each day of the diagnosis corresponds to a vector. It is in conformity to the clinical practice that the treatment for inpatient is on a daily basis. 
	
	\paragraph{Next day activity prediction} Table \ref{tab:recall_results} shows the results for next day activity prediction. Among the three approaches with original input, Dipole and T-LSTM achieve better performance than Retain, which has a trade-off between precision accuracy and interpretability. By inputting pre-train representations to the approaches with fine-tuning procedure, the prediction performance have changed in different scales. For Med2Vec, the change is small. This is because Retain, Dipole and T-LSTM adopt the same pooling strategy as Med2Vec for processing the original one-hot input. It means that the fine-tuning procedure is same as the end-to-end training procedure of the three prediction models. Therefore, when the models achieve convergence, the different initializations have limited impact on the final performance. However, the architectures of RoMCP and Inpatient2Vec can extract more proper representations for prediction with the help of fine-tuning procedure. In contrast, Inpatient2Vec contributes more to the prediction models, because the Transformer has a ability to handle various dependencies in inpatient data. 
	
	\paragraph{Remaining LOS prediction} The performance of remaining LOS prediction is shown in Table \ref{tab:recall_results}. We can observe that the approaches with Inpatient2Vec outperform others, and Med2Vec makes minimum contributes to the prediction models. The reasons are similar to next day activity prediction task. It is worth mentioning that even best performance achieved by Inpatient+T-LSTM is only 3.3084, which we can infer that the remaining LOS prediction is a difficult task. More information, such as lab testing and medical notes, should be introduced. 
	
	%\begin{figure}[htb]
	%	\centering
	%	\includegraphics[width=0.8\columnwidth]{pic/ablation}
	%	\caption{The pre-training model }
	%	\label{representation results}
	%\end{figure}
	
	\begin{figure}[htb]
		\centering
		\includegraphics[width=1.0\columnwidth]{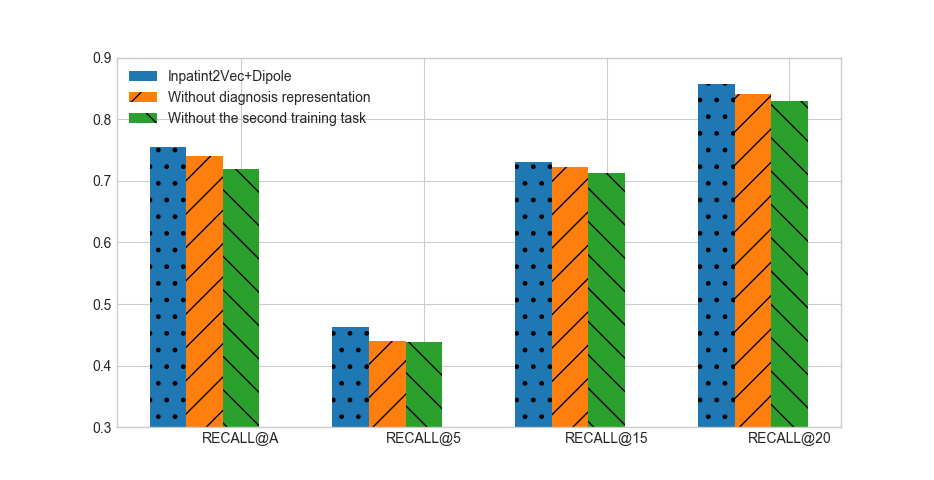}
		\caption{The results ablation study. }
		\label{abalation}
	\end{figure}
	
	\paragraph{Ablation Study} To evaluate the effectiveness of different components in Inpatient2Vec, we designed two comparative approaches. One is to remove our diagnosis representation, and treat each diagnosis as an activity. It is used to verify the importance of the proposed diagnosis representation for inpatient RL. The other one is to replace our second training task (next day activity prediction) by a pair-wise day prediction that given any two days, decides if they are consecutive. It is similar to the original training task (next sentence prediction) in BERT. This task focuses on testing the contribution of temporal relations for Inpatient2Vec. We use Dipole as the evaluation basic model to calculate the Recall@K for ablation study. Fig.\ref{abalation} shows the comparison results. We can make a conclusion that the two removed components are of great importance for inpatient RL.  
	
	\section{Conclusion}
	In this paper, we propose a novel Inpatient2Vec model to learn representations for inpatients. According to the distinctive data characteristics of inpatient, three kinds of representations, including activity, day and diagnosis, are combined by a Transformer-based network. The guidance role of the diagnosis and the dependency between unordered activities are well-designed in the network by a self-attention mechanism. We present two tasks, respectively focus on activity co-occurrence and day temporality, to train the networks. On a real-world dataset, semantic similarity measurement and inpatient clinical prediction are used as the evaluation tasks. The former one demonstrates that the learned activity and diagnosis representations can capture the clinical semantic information. The latter one shows the applicability of our method for prediction tasks by a fine-tuning procedure. 
	
	One limitation of the work is the model interpretability, which is necessary in clinical scenario. We will further study on it by incorporating domain knowledge or proper attention machanism. Another important future work is to integrate inpatient and outpatient data for a more comprehensive RL. 
	
	\section*{Acknowledgment}
	This research was supported by the National Key Technology Support Program (No. 2015BAH14F02).
	
	\bibliographystyle{IEEEtran}
	\bibliography{bibm2019}

\end{document}